\documentclass[runningheads]{llncs}
\usepackage[T1]{fontenc}
\usepackage{graphicx,verbatim}
\usepackage{textcomp}
\usepackage{stfloats}
\usepackage{url}
\usepackage{verbatim}
\usepackage{graphicx}
\usepackage{cite}
\usepackage{subcaption}
\usepackage{booktabs}
\usepackage{multirow}
\usepackage{amsmath,amsfonts,amssymb}
\usepackage{hyperref}
\usepackage{url}

\usepackage{xcolor}
\begin{document}
\title{Towards Fine-Grained and Verifiable Concept Bottleneck Models}
% \titlerunning{GenCBM}
\begin{comment}  %% Removed for anonymized MICCAI submission
\author{First Author\inst{1}\orcidID{0000-1111-2222-3333} \and
Second Author\inst{2,3}\orcidID{1111-2222-3333-4444} \and
Third Author\inst{3}\orcidID{2222--3333-4444-5555}}
\institute{Princeton University, Princeton NJ 08544, USA \and
Springer Heidelberg, Tiergartenstr. 17, 69121 Heidelberg, Germany
\email{lncs@springer.com}\\
\url{http://www.springer.com/gp/computer-science/lncs} \and
ABC Institute, Rupert-Karls-University Heidelberg, Heidelberg, Germany\\
\email{\{abc,lncs\}@uni-heidelberg.de}}

\end{comment}

\author{Yingying~Fang\inst{1}$^*$,
Haijie~Xu\inst{1}$^*$,
Shuang~Wu\inst{2},
Mariathasan~Anish\inst{1},
Guang~Yang\inst{1}}
\authorrunning{Y. Fang, H. Xu et al.}
\institute{Bioengineering Department and Imperial-X, Imperial College London, London, UK \and
    Thoughtworks AI Labs, Singapore \\
    \email{y.fang@imperial.ac.uk}}
  
\maketitle             

\renewcommand{\thefootnote}{}
\footnotetext{$^*$Equal contribution}
\renewcommand{\thefootnote}{\arabic{footnote}}

\begin{abstract}
Concept Bottleneck Models (CBMs) offer interpretable alternatives to black-box predictors by introducing human-relatable concepts before the final output. However, existing CBMs struggle to verify whether predicted concepts correspond to the correct visual evidence, limiting their reliability. We propose a fine-grained CBM framework that grounds each concept in localized visual evidence, enabling direct inspection of where and how concepts are encoded. This design allows users to interpret predictions and verify that the model learns intended concepts rather than spurious correlations. Experiments on medical imaging benchmarks show that our learned concept space is information-complete and achieves predictive performance comparable to standard CBMs, while substantially improving transparency. Unlike post-hoc attribution methods, our framework validates both the presence and correctness of concept representations, bridging interpretability with verifiability. Our approach enhances the trustworthiness of CBMs and establishes a principled mechanism for human–model interaction at the concept level, paving the way toward more reliable and clinically actionable concept-based learning systems. Our code is released at \url{https://anonymous.4open.science/r/Concept_Grounding-4FF8/}. 

% Authors must provide keywords and are not allowed to remove this Keyword section.

\end{abstract}
\section{Introduction}
The emergence of Concept Bottleneck Models (CBMs) \cite{koh2020concept} marks a paradigm shift from post-hoc attribution toward inherently interpretable learning. Unlike conventional black-box models, CBMs explicitly introduce an intermediate reasoning layer composed of interpretable concepts before producing the final predictions. By constraining predictions to pass through this semantic bottleneck, CBMs aim to ensure that model reasoning aligns with domain-relevant concepts that are interpretable by humans~\cite{huy2025interactive,barnett2021case}.

A seminal step toward concept-based interpretability was TCAV~\cite{kim2018interpretability}, which quantified the sensitivity of learned representations to predefined concepts. Building on this idea, Koh et al.~\cite{koh2020concept} formalized the CBM architecture by introducing an explicit concept layer. Early CBM research primarily focused on maintaining predictive performance, \emph{i.e.} ensuring that the extracted concepts retained the full discriminative power of the original features. To this end, some works explored architectural modifications such as autoregressive models~\cite{havasi2022addressing} and transformer-based designs~\cite{gao2024aligning,wang2025mvp}. Others sought to improve the information completeness of the concept space by augmenting explicit concept representations with implicit feature information~\cite{espinosa2022concept,sheth2023auxiliary,zhang2024decoupling}. In parallel, another line of research aimed to reduce reliance on fully supervised concept annotations, leveraging weak supervision or text-guided signals to scale CBMs more broadly~\cite{oikarinen2023clip,oikarinen2023label,tan2024explain}.

Despite these advances, a critical limitation remains: the internal visual grounding of predicted concepts is often opaque. While CBMs expose concept activations, they do not inherently verify whether those concepts correspond to the correct visual evidence in the input image. To address this issue, recent works have explored concept grounding via visualization techniques such as Grad-CAM, cross-alignment, or human-annotated supervision~\cite{huang2024concept,xie2025discovering,chanda2024dermatologist}. 
% For example, 
ProtoCBM~\cite{huang2024concept} aligns heatmaps across layers and augmented samples to improve localization consistency, while DOT-CBM~\cite{xie2025discovering} leverages vision–text alignment to identify concept-relevant regions. Derma~\cite{chanda2024dermatologist} supervises grounding maps using human-provided localization annotations.

However, these approaches still face notable limitations. Grad-CAM\cite{selvaraju2017grad,chattopadhay2018grad} typically produces coarse and spatially diffuse localizations, which are particularly problematic in medical images where discriminative cues are subtle and anatomically constrained. Cross-alignment methods attempt patch-level correspondence, but they rely on pretrained multimodal encoders with fine-grained vision–text alignment, the training of which is itself a complex and resource-intensive research problem.
Methods that depend on grounding annotations further incur substantial labeling costs and limit scalability to new concepts or datasets. 
Consequently, there remains a need for a fine-grained, reliable, and annotation-free grounding mechanism that can strengthen CBMs into fully verifiable interpretable models.

In this work, we propose a Generative Concept Bottleneck Model (GenCBM) that equips each concept with verifiable and fine-grained localization. Our central insight is that generative features, unlike purely discriminative ones, offer a complete latent space better suited for concept learning. Training CBMs on generative features also allows robust tracing of concepts back to specific latent factors through manipulation of concept-related directions in the latent space and counterfactual reconstructions. Concept grounding is then achieved through contrastive visualization between reconstructed and counterfactual images, enabling direct verification and better reliability.

Our contributions are summarized as follows:
(1) We propose a novel CBM framework premised on generative features, offering a principled alternative to conventional discriminative backbones.
(2) We demonstrate that generative CBMs integrate seamlessly with counterfactual grounding mechanisms, enabling fine-grained concept-level tracing and verification. This substantially improves the trustworthiness of concept-based explanations by ensuring that concepts are both interpretable and verifiable.
(3) Extensive experiments on medical imaging benchmarks demonstrate that our method achieves state-of-the-art concept prediction accuracy together with more precise and reliable grounding results, underscoring its practical value in high-stakes clinical applications.

% \section{Related work}

% \rd{Different kinds of CBMs}

% \rd{++Introduce works to add localization ability of CBM: GradCAM, AAAI.}

% \rd{Highlight these localization methods}

\section{Method}

\begin{figure}[h]
\begin{center}
\includegraphics[width=1\linewidth]
{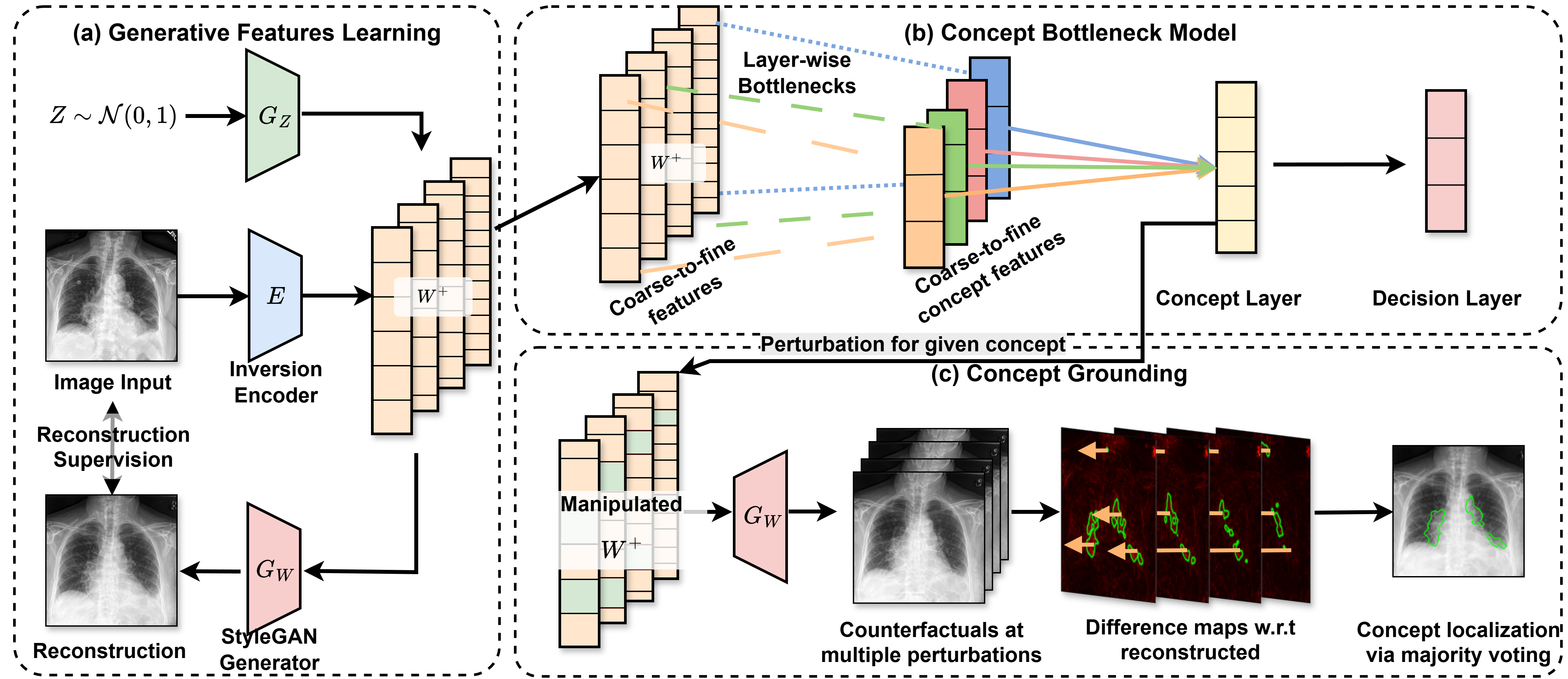}
\end{center}
\caption{
Framework of \textbf{GenCBM}. Left: We learn generative features by training a StyleGAN generator along with a latent inversion encoder. Right top: The coarse-to-fine generative features are used for concept learning in the CBM stage. Right bottom: We generate counterfactuals by perturbing generative features along the concept activation vector. The difference maps with respect to the unperturbed reconstruction are aggregated to localize the concept activation.}\vspace{-2em}
\label{fig:framework}
\end{figure}

\subsection{Overview}
As illustrated in Fig.~\ref{fig:framework}, a key component of our approach is the use of StyleGAN latents as the base feature representations for concept learning. By operating in this generative latent space, concept activations can be directly localized in the input image through counterfactual generations and difference maps. This concept grounding is more robust and less prone to noise disturbance and spurious correlations. In what follows, we elaborate on the key components and design rationales in our framework.

\subsection{Generative Features}
Previous CBMs typically extract concepts using discriminative feature backbones such as ResNet~\cite{he2016deep}. A key design distinction in our approach is to instead build CBMs on top of generative features. This choice offers two key benefits. \textbf{Information Completeness}: Generative features preserve the full spectrum of information required for concept learning, ensuring that rare or subtle concepts arising from long-tail events are retained, whereas discriminative representations often underrepresent or omit them. \textbf{Robustness and Stability}: Gradient-based saliency methods~\cite{selvaraju2017grad,chattopadhay2018grad} emphasize instance-level pixel variations, a perspective that frequently yields unstable and noisy highlights. In contrast, visualizing concept activations through generative features inherently entails inspecting latent encodings across the entire data distribution, resulting in consistent, semantically aligned visualizations and reducing susceptibility to spurious correlations. To this end, we propose our GenCBM framework to integrate generative features and concept bottlneck models.

Specifically, we train a StyleGAN~\cite{karras2020analyzing} generator $G$. The choice of StyleGAN is motivated by its disentangled latent codes, facilitated by adaptive instance normalization during training, which is particularly advantageous for concept learning. We also train an encoder $E$ using the pSp framework~\cite{richardson2021encoding} to maps input images into latent codes in the $\mathcal{W}^+$ space. These StyleGAN latent codes, serve as the generative features upon which our concept bottleneck is constructed. By leveraging this generative latent space, our approach ensures richer information coverage, improved disentanglement of concepts, and more stable visualizations

Formally, the pre-trained StyleGAN generator can be seen as a composition of two networks $G=G_w\circ G_z, G_z:z\mapsto \mathcal{W}^+ $, where $z$ is sampled from a Gaussian distribution. The encoder $E$ encodes a given input image $I$ as $\mathbf{w}^{+} = E(I)\in \mathcal{W}^+$, which can then be mapped back into image space via $I_{\text{Rec}} = G_w(\mathbf{w}^{+}).$ We train this encoder $E$ via the following reconstruction objective\footnote{The generator $G_W$ is jointly fine-tuned during this process.}:
\begin{equation}
\mathcal{L}_{\text{rec}}
= \lambda_{\text{pixel}}\|I_{\text{rec}} - I\|_{1}
+ \lambda_{\text{perceptual}}\|\phi(I_{\text{Rec}}) - \phi(I)\|_{2}^{2}
+ \lambda_{w}\|\mathbf{w}^{+} - \bar{\mathbf{w}}^{+}\|_{2}^{2}.
\end{equation}
Here, the perceptual term employs a fixed feature extractor $\phi(\cdot)$ (e.g. VGG~\cite{simonyan2014very}) and the regularization term minimizes the deviation of $\mathbf{w}^+$ from its mean $\bar{\mathbf{w}}^{+}$.

\subsection{Hierarchical $\mathcal{W}^+$ Concept Bottleneck Training}
The StyleGAN latent $\mathbf{w}^{+} \in \mathbb{R}^{N\times L}$ comprises $L$ layers, each of dimension $N$. By design, these layers exhibit a hierarchical, coarse-to-fine control over image attributes, ranging from global geometry to mid-level structure and local texture~\cite{karras2020analyzing,richardson2021encoding,xu2021generative}. 
To account for this hierarchical structure of the $\mathcal{W}^+$ space in our concept bottleneck model, we first perform intra-layer bottleneck via linear projections to extract \emph{layer-wise concept features}. These concept features preserve the disentangled structure of the $\mathcal{W}^+$ space and encode coarse-to-fine attributes. The overall concept predictions are obtained via an aggregated linear projection of these coarse-to-fine concept features. This design ensures that concept learning respects the multi-scale structure of the latent space, capturing both global and fine-grained visual evidence.

\subsection{Progressive Concept Grounding via Counterfactuals}
To localize the concept region in the original image, we generate counterfactual images by enhancing or suppressing the corresponding concept features, followed by calculating the difference map between the original and counterfactual images. However, the perturbation magnitude is critical: an excessively large manipulation may amplify the target concept while also altering correlated features, whereas an overly small manipulation may produce insufficient visual changes for reliable localization.
To explicitly ground the learned concepts, we propose a progressive manipulation strategy. Specifically, we perturb the latent representation $\mathbf{w}^+$ along the concept activation direction using a predefined spectrum of perturbation magnitudes, thereby generating a sequence of counterfactual images with gradually increasing concept changes. This progressive strategy enables concept localization from fine to coarse. By comparing the difference maps across perturbation levels and identifying regions that consistently emerge at early stages and remain spatially overlapping across larger perturbations, we obtain the earliest and most relevant fine-grained regions as the grounding evidence for each concept.

For a counterfactual image $I_e$ with perturbation $e$, and the reconstructed image $I_0$, we quantify their pixel-wise difference map $D_e$ by averaging across RGB channels $D_e(x, y) = \frac{1}{3} \sum_{c \in \{R,G,B\}} \left| I_e^c(x, y) - I_0^c(x, y) \right|$.
% Compared to grayscale differencing, this is more effective at capturing chromatic shifts at equal luminance, such as variations in pigmentation or erythema distribution, which are of particular clinical significance in dermoscopic analysis.
To mitigate local noise and normalize scale differences across perturbation levels, we apply a Gaussian filter ($\sigma = 3$) to obtain a smoothed map $\tilde{D}_e$, followed by an adaptive binarization strategy. The binary mask $M_e$ is derived using a dynamic threshold:
$\tau_e = \max(\text{Percentile}_{95}(\tilde{D}_e), \delta_{\min})$, where the 95th percentile extracts the most salient regions, and the lower bound $\delta_{\min} = 5$ on a 0--255 UINT8 scale filters out background noise at low perturbation levels. Typically, masks for different $|e|$ exhibit a trade-off between specificity in the localized concept activations and noise ($\downarrow |e| \implies \uparrow \text{specificity} \& \downarrow \text{noise}$). We aggregate these difference masks via a pixel-wise majority voting mechanism $M_{\text{vote}}(x,y) = \mathbf{1}\left[\sum_{e \in \mathcal{E}} M_e(x,y) \geq \theta\right]$, with $\theta$ set to 5 empirically. This voting ensures both consistency and coverage in the final concept localization prediction.

% The difference masks $M_e$ generated at smaller $|e|$ exhibit minimal noise but yield broader regions of interest, whereas larger $|e|$ produce finer, highly localized foci at the cost of increased noise.

% \paragraph{Mask Aggregation and Post-processing}
% Masks generated at smaller $|e|$ exhibit minimal noise but yield broader regions of interest, whereas larger $|e|$ produce finer, highly localized foci at the cost of increased external noise. To synergize these complementary properties, we employ a pixel-wise majority voting mechanism. The aggregated mask is formulated as:
% $$M_{\text{vote}}(x,y) = \mathbf{1}\left[\sum_{e \in \mathcal{E}} M_e(x,y) \geq \theta\right]$$
% with the voting threshold $\theta$ empirically set to 5. 

% Finally, to refine the spatial coherence of $M_{\text{vote}}$, we sequentially apply morphological closing and opening operations (both with 2 iterations) to fill internal holes, bridge fragmented segments, and eliminate isolated noise patches. We then perform 4-connected component labeling and retain only the top $K=3$ largest components to discard residual debris, ultimately yielding the final concept saliency mask, which is visualized as overlaid contours on the reconstructed image.

\section{Experimental Details}

\noindent\textbf{Datasets.}
To evaluate the efficacy of our GenCBM framework, we conducted experiments across two imaging modalities: dermoscopy and chest X-ray. For dermoscopy, we formulated a binary classification task to separate melanoma from non-melanoma using 3,611 images from the ISIC2018 archive~\cite{codella2019skin,tschandl2018ham10000}\footnote{\url{https://challenge.isic-archive.com/data/\#2018}}, enriched with seven melanoma-related concepts and grounding annotations~\cite{chanda2024dermatologist}. An image is classified as exhibiting melanoma if two or more such concepts are predicted. For chest X-ray, we framed a lung abnormality detection task using 224,316 radiographs from the CheXpert dataset~\cite{irvin2019chexpert}\footnote{\url{https://stanfordmlgroup.github.io/competitions/chexpert/}}. Following the dataset’s 14 abnormality concepts, an image is classified as abnormal if at least one is detected.
% We constructed a balanced evaluation set containing \bl{20,000 normal cases (no abnormality) and 20,000 abnormal cases}.

% \subsection{{Experimental Details}}
% \subsection{Implementation Details and Comparison Baselines}
\noindent\textbf{Implementation Details and Comparison Baselines.}
For both datasets, input images were resized to $256 \times 256$. Generative feature representations were obtained by training a StyleGAN generator for 100 epochs, followed by pSp encoder training (both optimized with Adam\cite{kingma2014adam} at a learning rate of $1\times10^{-4}$ and batch size of 64), yielding a feature representation $\mathbf{w}^+\in\mathbb{R}^{512\times14}$. Our GenCBM were trained with similar learning rate and batch size, with each layer-wise concept feature dimension set to 32 for extracting the hierarchical concept features from $\mathbf{w}^+$.
Our GenCBM is benchmarked against representative baselines, including the vanilla CBM~\cite{koh2020concept}, the Transformer-based variant Explicd~\cite{gao2024aligning}, and recent CBMs focusing on concept grounding (ProtoCBM~\cite{huang2024concept}) and concept–image alignment (DOT-CBM~\cite{xie2025discovering}). For ISIC, we additionally include Derma~\cite{chanda2024dermatologist}, which leverages concept grounding annotations for supervision. All baselines were fine-tuned using their official implementations on ISIC and CheXpert. For concept localization and visualization, we applied Grad-CAM~\cite{chattopadhay2018grad} to vanilla CBM and ProtoCBM and used text–vision alignment for Explicd and DOT-CBM.

\section{Experimental Results}
\begin{figure}[h]
    \centering
    % Left Image
    % \subfloat[ISIC \rd{Font enlarged, decision using another color, C0-F1}]
    {\includegraphics[width=0.45\textwidth]{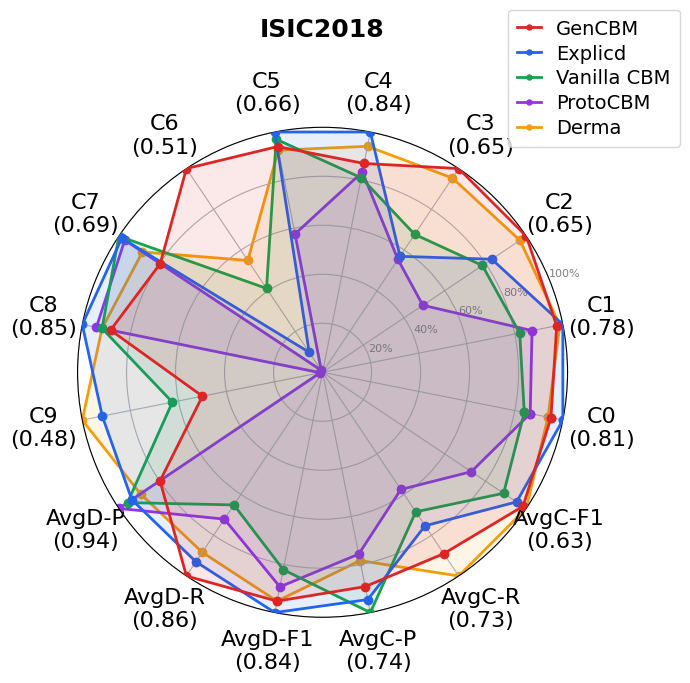}\label{fig:sub1}}
    \hfill 
    % Right Image
    % \subfloat[CheXpert]
    {\includegraphics[width=0.45\textwidth]{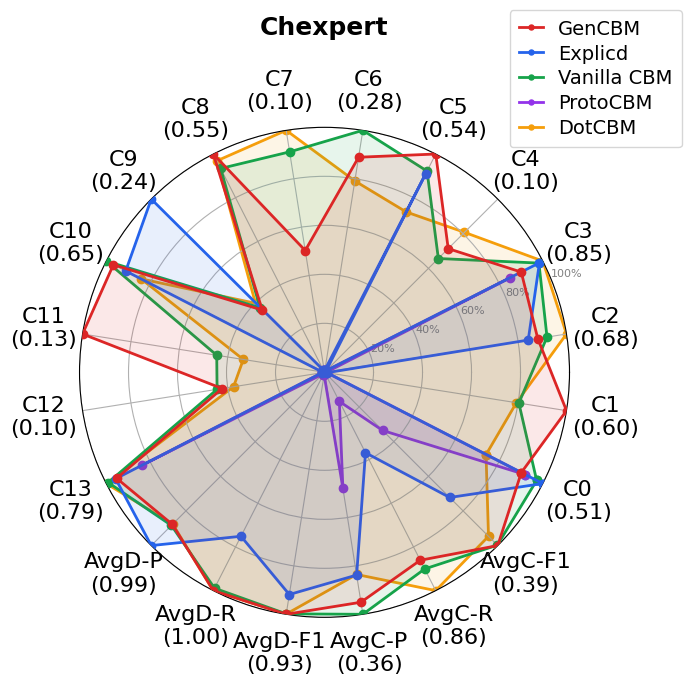}\label{fig:sub2}}
    \caption{Concept prediction performance across different methods. We report the F1 score for each concept and the precision/recall/F1 for the overall average.}\vspace{-2em}
    \label{fig:concept_accuracy}
\end{figure}

\subsubsection{Concept Prediction Performance.} 
As shown in Fig.~\ref{fig:concept_accuracy}, our method delivers competitive concept prediction performance compared to state-of-the-art CBM baselines on both ISIC and CheXpert. 
% Result one
On ISIC, our GenCBM consistently rank among the top performing methods for the F1 score across most concepts, along with Derma~\cite{chanda2024dermatologist} and  Explicd~\cite{gao2024aligning}. Our method also achieves the best overall F1 score (averaged over all concepts) at 63\%, along with Derma (63\%) and edging out Explicd (60\%).
% Result two
On CheXpert, GenCBM (39\%), Vanilla CBM (39\%), and DotCBM (37\%) achieve the highest average F1 scores. Across both datasets, these results confirm that our method delivers concept prediction accuracy on par with state-of-the-art CBMs built on discriminative features. In the next subsection, we turn to its more critical advantage—producing robust and fine-grained concept localizations.

\begin{figure}[h]
    \centering
    % Change width to 1.0\textwidth to fill the horizontal space
    \includegraphics[width=0.9\textwidth]{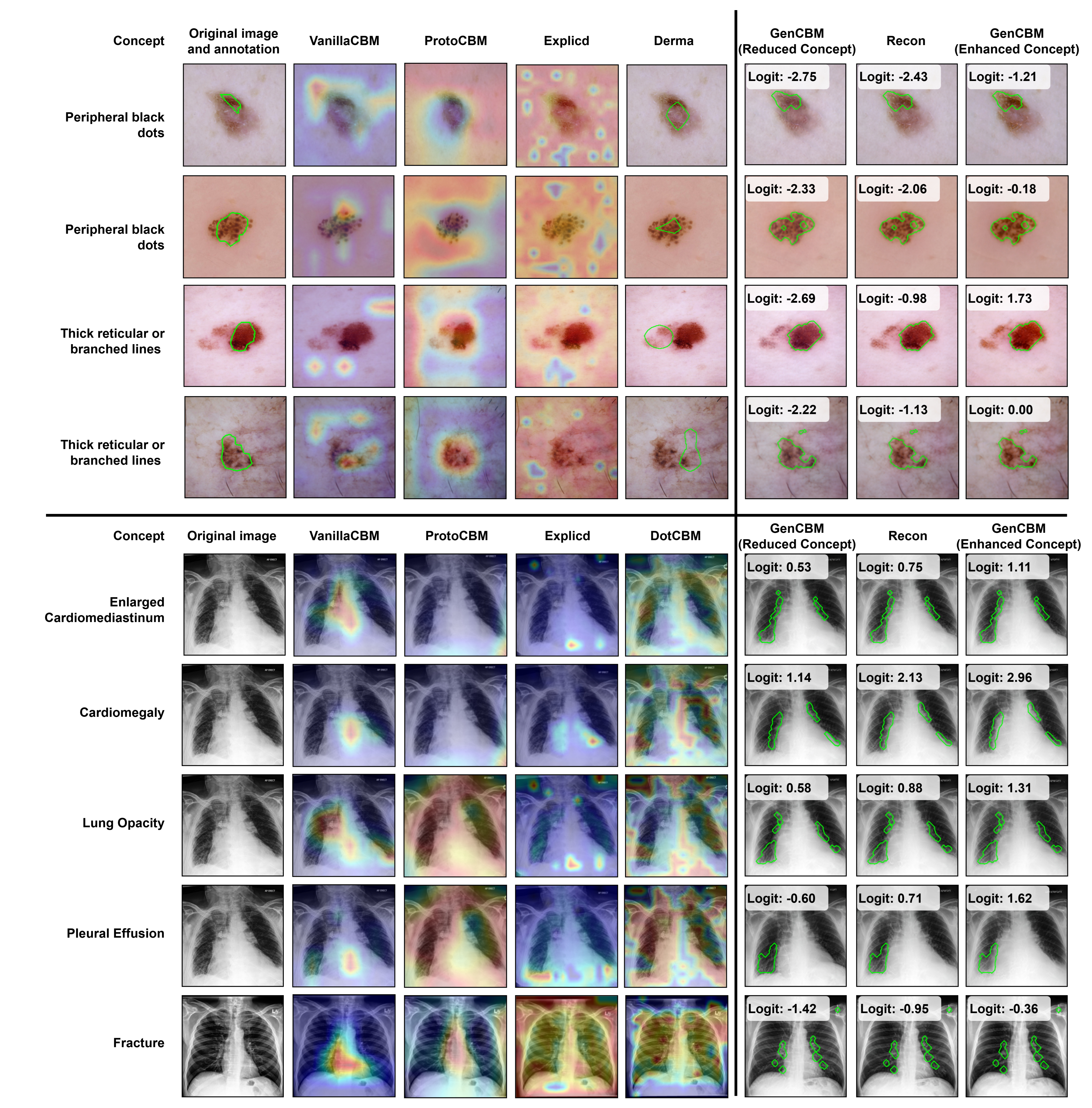} 
    \caption{Concept grounding visualizations produced by different CBMs. On ISIC, we show localization maps for the same concept across different images, highlighting the consistency of localization patterns across models. On chest X-ray, we show localization maps for different concepts within the same image, demonstrating each model’s ability to localize multiple abnormalities in a single case. For GenCBM, we also show the logits values for the counterfactuals at $e=\pm20$.}\vspace{-2em}
    \label{fig:single_visualization}
\end{figure}

\subsubsection{Concept Grounding Evaluation.}
Fig.~\ref{fig:single_visualization} illustrates concept localization results for a subset of representative concepts. Across all cases, GradCAM visualizations from Vanilla-CBM and DotCBM yield coarse, overly broad localizations that lack specificity. ProtoCBM and Explicd produce more focused regions, but often capture background noise or spurious features. Derma achieves improved lesion-related localization on ISIC, which can be attributed to its use of localization supervision during training. In contrast, our method provides the most accurate and fine-grained grounding, notably without requiring any additional localization annotations.  Moreover, it exhibits consistent localization patterns across diverse images, supporting our hypothesis that concept grounding via generative features and counterfactual generation is more robust. For Chexpert, prior methods show overlapping coarse regions with little concept specificity, whereas our GenCBM successfully identifies distinct, concept-specific anatomical features. Overall, the ability to deliver fine-grained, anatomically precise localization of medical concepts is the defining strength of our approach.
% By coupling interpretable concepts with their corresponding visual activations, our method empowers end users to both understand the model’s reasoning and validate it against observable evidence.
\begin{figure}[h]
    \centering
    % --- 第一行：原有的三张图 ---
    % Left Image
    {\includegraphics[width=0.3\textwidth]{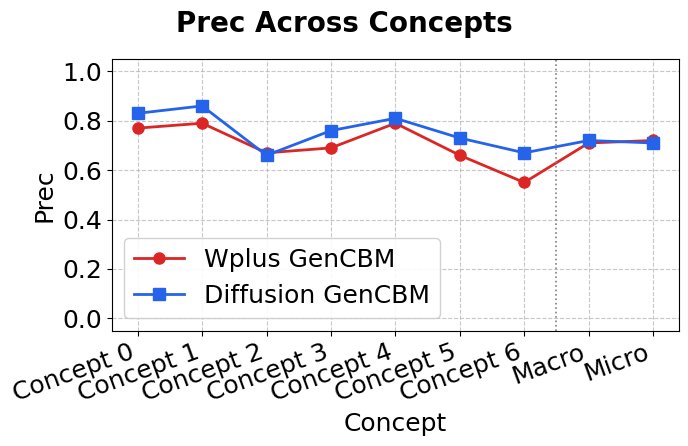}\label{fig:diff2_A}}
    \hfill
    % Middle Image
    {\includegraphics[width=0.3\textwidth]{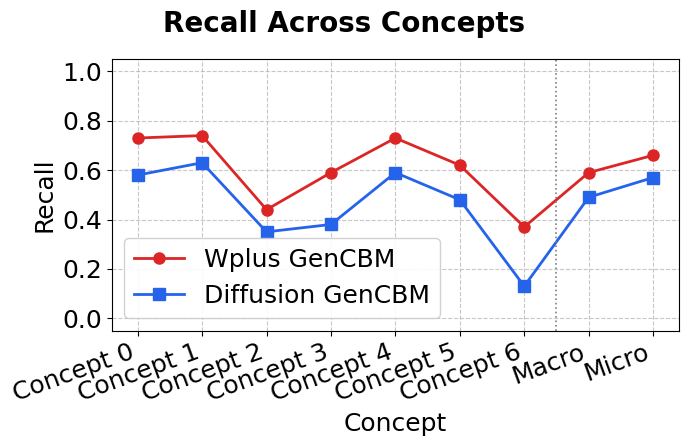}\label{fig:diff2_B}}
    \hfill
    % Right Image
    {\includegraphics[width=0.3\textwidth]{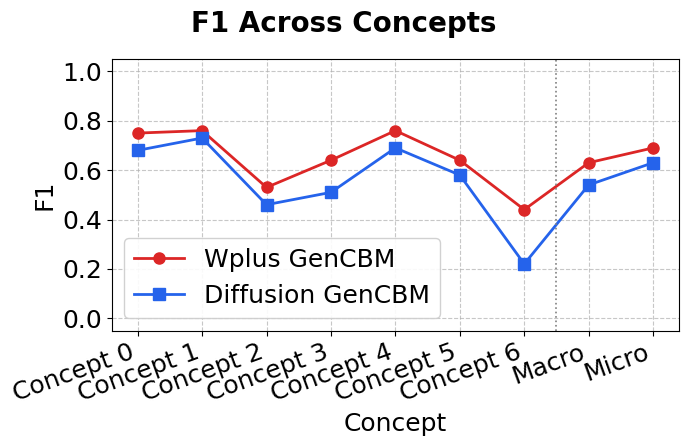}\label{fig:diff2_C}}
    \caption{Concept prediction performance of GenCBM under different representation spaces. 
    % The W+ space achieves significantly higher recall and consequently a higher F1 score on ISIC.
    }
    \vspace{-2em}
    \label{fig:representation}
\end{figure}
\subsubsection{Ablation study.}
We conducted an ablation study on the choice of generative features by comparing StyleGAN latents (W+) with diffusion-based autoencoder representations~\cite{preechakul2022diffusion} on ISIC. While diffusion models are well known for strong image reconstruction and generation, their reconstruction relies on both the conditioning signal and injected noise. This motivates our hypothesis that representations derived solely from the conditioning pathway may omit discriminative information essential for concept learning. Our empirical findings support this hypothesis. As shown in Fig.~\ref{fig:representation}, adopting StyleGAN W+ latents within GenCBM consistently yields superior concept prediction performance compared to diffusion-based latents.

\section{Conclusion}
We demonstrate that integrating concept bottleneck models with generative features provides a compelling alternative to conventional CBMs, yielding significant improvements in robust and faithful concept grounding. This empowers end users not only to engage in the interpretability offered by CBMs but also to validate those concepts via their visual activations. By coupling interpretability with verifiability, GenCBM represents a critical advancement toward trustworthy and clinically meaningful decision support.

\clearpage
\bibliographystyle{splncs04}
\bibliography{reference}
\end{document}